# ProtoSemImage: Image-Valued Prototypes with Deformable Row Alignment for Interpretable Document Classification

Mohammad Zare
Artificial Intelligence Lab at
AriooBarzan Engineering Team
Shiraz, Iran
md.zare@sutech.ac.ir

Pirooz Shamsinejadbabaki
Department of Computer Engineering
and Information Technology
Shiraz University of Technology,
Shiraz, Iran
p.shamsinejad@sutech.ac.ir

***Abstract*— Prototypes in classification models are almost always vectors, and a vector has no readable form. This paper asks what happens when a prototype is an image. Documents give the question a natural form, because a document can be rendered as a multi-channel image in which every token becomes a pixel, so a class representative can take the same shape and the same channel semantics as the inputs it stands for. ProtoSemImage represents each class by one or more visual archetypes: prototype images in a four-channel HSV space whose channels carry named linguistic factors. A Skip-Gram objective learns that color space end to end through a four-dimensional bottleneck, discourse boundary rows become differentiable typed difference rows, and classification reduces to 2D visual template matching: a deformable row alignment between a document image and the archetype bank, in the spirit of dynamic time warping. Because the match is a spatial pattern comparison rather than a linear readout, the model reports where an input departs from its archetype and along which channel, and a generative head decodes each archetype back into text. The image representation works: it beats an otherwise identical model with vector prototypes in all three paired seeds, by between 4.3 and 11.8 points on a ten-class task. The distance-based matching does not. A diagnostic that keeps the representation fixed and swaps only the classifier recovers the sequence baselines, which locates a 20.6-point shortfall in the matching rather than in the color compression, and a benchmark built so that a pair of documents shares a bag of words and differs only in arrangement confirms the layout-preservation it was designed for. We report both directions, because for a representation whose whole purpose is inspect ability, the failure modes are as informative as the gains.**

## I. Introduction

Prototype-based classifiers occupy a settled place in computer vision. A prototypical network represents each class by a point in embedding space, usually the mean of its members, and classifies by nearest distance [10]. Interpretable variants replace that abstract point with concrete image patches, so a prediction can be justified by showing which training evidence it resembles [11]. Both families rest on an assumption that has gone largely unexamined. The prototype is a vector, or at most a small patch, and neither can be looked at as a whole. A mean vector is also a bag: it discards the arrangement of its members entirely, so two inputs holding the same content in a different order produce the same prototype.

This paper asks what happens when the prototype is a full image. Documents give the question a natural form, because a document can be rendered as a multi-channel image in which every token becomes a pixel. A class representative can then be an image of the same shape and the same channel semantics as the documents it stands for. Classification becomes 2D visual template matching against a small set of such templates, and the templates can be displayed.

Documents have been rendered as images before. SemImage [9] draws each sentence as a row of pixels, inserts boundary rows whose brightness tracks semantic dissimilarity, and assigns meaning to color: hue for topic, saturation for sentiment, value for emphasis. The design buys transparency at the input, and an ablation shows the channel assignment does real work. Three limits remain. The classifier on top is an ordinary softmax, so the decision stays opaque. Channel separation rests on auxiliary losses, a soft constraint that degrades once they are removed. And the boundary rows come from a frozen sentence encoder, fixing the structure the network sees before training begins.

ProtoSemImage answers that question in the affirmative, with a caveat the experiments make plain. Every class is represented by a small set of visual archetypes, each one a prototype image in the same four-channel HSV space used to render documents. Nothing about the arrangement is decorative. An archetype occupies the same grid as a document, so unlike a point it is at least capable of encoding discourse order: in principle an archetype could record that documents of its class open on a neutral note and turn negative partway through. Whether a trained model actually exploits that capacity is a separate question from whether the representation permits it, and Section VI reports what we found when we tested it. Being an image, the archetype can also be displayed. Sharing the channel semantics of the document space, it yields a typed and localized account of disagreement rather than a single scalar. The color space is trained under a Skip-Gram objective, so four channels form a bottleneck that must predict context words, which grounds the space in distributional statistics and lets boundary rows be learned instead of inherited from a frozen encoder. A generative head maps archetypes back to token distributions, so an archetype can be read as well as viewed.

- We introduce image-valued prototypes: a class is represented by one or more learned prototype images in a disentangled color space, which preserves document layout and makes the prototype an inspectable object.

- We learn that color space end to end under a Skip-Gram objective over a four-dimensional HSV bottleneck, which grounds channel semantics in corpus statistics and makes discourse boundary rows differentiable rather than inherited from a frozen encoder.

- We adapt soft dynamic time warping to row-axis alignment between document images and archetypes, handling variable length and returning the correspondence path used for explanation.

- We isolate what the image representation contributes with an otherwise identical vector-prototype model, and locate the remaining accuracy shortfall with a diagnostic that swaps only the classifier.

Section II reviews related work, Section III develops the framework, Sections IV and V report results and ablations, and Sections VI and VII set out what the evidence supports and where the approach falls short..

## II. RELATED WORK

### A. Text as an Image

Character-level and word-level convolutional models established that a document can be read from a one-dimensional pixel strip [12], [13], and character-level transformers softened tokenization further [14], but all of them keep the document one-dimensional and so limit the network to patterns running along a single axis. Transformer encoders [1], [2], [3] now set the accuracy standard on this task [4], though their decisions resist inspection. SemImage [9] broke with that arrangement by turning the document into a spatial pattern a convolutional network can read, with a color space whose channels carry named linguistic factors, with boundary rows marking semantic discontinuities. ProtoSemImage keeps the two-dimensional layout and the channel semantics, then replaces the softmax head with distance-based archetype matching and the fixed boundary computation with a learned one.

### B. Disentangled Representations and Prototypes

Variational objectives with a strengthened independence penalty became a standard tool for separating factors of variation [6], and information-theoretic variants extended the idea to discrete factors [7]. A sobering result showed that unsupervised disentanglement is impossible without inductive bias or supervision [15], which shapes the design here: rather than hoping a penalty term will separate topic from sentiment, ProtoSemImage gives each factor a physical channel and compares channel by channel, which enforces separation structurally rather than through the shared parameters of multi-task learning [5], [8]. Prototypical networks classify by distance to a class mean [10], and interpretable variants tie each prototype to concrete evidence [11], but both assume the prototype is a vector, so order is discarded and the prototype cannot be read. ProtoSemImage relaxes that assumption: a prototype becomes an array with the same shape and channel semantics as a document image.

### C. Discourse Structure and Alignment

Detecting where a document changes subject has traditionally relied on lexical cohesion [18], and sentence encoders trained on paraphrase data made similarity-based segmentation practical [19]. Such methods compute boundaries in a separate pass and hand the result downstream as fixed input. ProtoSemImage folds the computation into the network. Dynamic time warping has aligned variable-length sequences since its introduction for speech recognition [20], and its soft relaxation made the recursion differentiable [21]. Here that relaxation runs along the row axis, and the resulting path becomes the correspondence map the explanation is built on. Self-attention [16] is the alternative that has largely displaced such recursions, at the cost of the explicit correspondence an alignment path provides.

## III. METHODOLOGY

ProtoSemImage turns a document into an image, learns a color space in which that image is meaningful, and classifies by comparing it against a bank of archetypes. The pipeline appears in Figure 1.

### A. End-to-End Continuous HSV Bottleneck

Every token $w_{i,j}$ is first mapped to a learnable embedding of dimension d, drawn from a table E trained jointly with the model rather than taken from a frozen encoder:

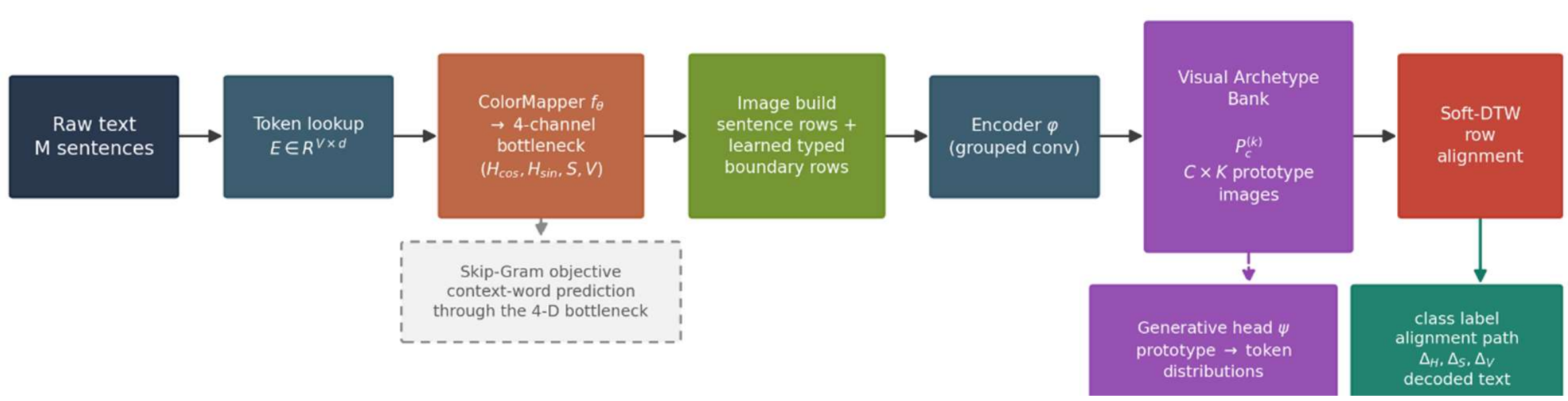

*Figure 1. The ProtoSemImage pipeline. Tokens pass through a four-channel HSV bottleneck trained by a Skip-Gram objective, become a document image with learned typed boundary rows, and are matched against a bank of archetype images by soft-DTW row alignment.*

$$e_{i,j} = E[w_{i,j}], EinR^{Vxd} (1)$$

A small network $f_{the}$ then projects that embedding into four channels. Two carry hue as unconstrained 2D Cartesian coordinates, each bounded independently by tanh, which avoids the discontinuity a circular angle introduces at the wrap-around point without imposing the unit-circle constraint $Hcos^2 + Hsin^2 = 1$ ; only the direction of the pair is meaningful. Assigning a color to an input is a colorization problem, and learned colorization has an established line of work in this community [26]; the difference here is that color is assigned per token to carry a named linguistic factor rather than to reproduce a reference image. Saturation and value are bounded by a sigmoid:

$$p_{i,j} = [Hcos, Hsin, S, V]^T = f_theta(e_{i,j}) (2)$$

$$Hcos = tanh(w_H c^T e + b_H c), Hsin = tanh(w_H s^T e + b_H s) (3)$$

$$S = sigma(w_S^T e + b_S), V = sigma(w_V^T e + b_V) (4)$$

The four dimensions are a bottleneck, and that is the point. Training drives the mapper with a Skip-Gram objective [23] in which a token's pixel vector must predict the words around it. A decoder g lifts the pixel vector back to dimension d for scoring against an output table U:

$$L_S G = -sum_{i,j} sum_{-c \leq k \leq c, k!=0} logp(w_{i,j+k} | w_{i,j}) (5)$$

$$p(w'|w) = exp\left(u_{w'}^T g(p_{i,j})\right) / sum_{w''} exp\left(u_{w''}^T g(p_{i,j})\right) (6)$$

The context window stays inside a sentence. Tokens near a sentence boundary attend only to neighbors on the same row, so the objective never mixes information across sentences and the row structure is left for the alignment stage to handle.

No other path connects a token to its context. Whatever distributional information the model uses to predict neighboring words has to survive compression into four numbers, and those four numbers are the pixel a person sees. Auxiliary losses on the pooled hue and saturation channels keep the assignment honest.

### *B. Dynamic Discourse Boundary Learning*

A sentence is summarized by the mean of its pixel row, Eq. 7:

$$s_i = (1/L) sum_{j=1}^{L} p_{i,j} (7)$$

Because $s_i$ is a function of the learned pixels, the boundary row built from it is differentiable. Earlier formulations used fixed Sentence-BERT embeddings [19] and wrote a single brightness value [9]. Boundary rows here carry a typed difference: the hue channels record the direction the topic moves, saturation the size of the sentiment shift, and value a learned discourse intensity:

$$B_i = [Hcos(s_{i+1}) - Hcos(s_i), Hsin(s_{i+1}) - Hsin(s_i)] \quad (8)$$

$$|S(s_{i+1}) - S(s_i)|, sigma(w_B^T [s_i; s_{i+1}])^T (9)$$

A reader can therefore distinguish a topic shift from a change in tone without consulting the model, which a single brightness value does not permit.

### *C. Document Image Construction*

Sentences and boundaries alternate, giving an image of height 2M-1:

$$I(D)[2i-1, j, :] = p_{i,j}, I(D)[2i, j, :] = B_i forj = 1..L (10)$$

The boundary vector is replicated across all L columns of its row, so a discourse break appears as a full-width horizontal band rather than a single point.

Documents are bucketed by sentence count during training, so no padding rows reach the encoder; padding would otherwise enter the correspondence path as spurious alignments.

### *D. Visual Archetype Bank*

Each class c owns K archetype images. Setting K above one is deliberate: a class whose members fall into several distinct modes is poorly served by a single representative, and averaging across modes yields a prototype resembling none of them. An archetype has the same shape and channel semantics as a document image:

$$P = P_c^{(k)} inR^{(2Mp-1)xLx4}, c = 1..C, k = 1..K (11)$$

Archetypes are not initialized at random: document images of each class are clustered in a reduced feature space, in the manner of deep clustering for representation learning [17], and the cluster centers seed them. A penalty term anchors each archetype to its seed, since an unconstrained image prototype drifts toward a constant array or toward noise that lowers the loss without representing anything.

### *E. Shared Encoder with Channel-Grouped Convolution*

One encoder phi process both documents and archetypes, which guarantees that the two live in a comparable feature space. The first convolution is grouped with four groups, one per input channel, so hue cannot leak into saturation at the earliest layer; a standard four-channel convolution mixes them immediately, which makes a channel-wise reading of the difference maps unreliable. The final convolution has kernel 3, stride 2 and padding 1, which on 2M-1 input rows yields exactly M output rows, one feature vector per sentence:

$$phi(I(D)) = F = [f_1, \ldots, f_M] (12)$$

Features are L2-normalised. Without that step the squared distances driving the alignment term scale with the feature width and swamp the classification objective.

### *F. Deformable Row Alignment*

Pixel-wise distance between a document image and an archetype fails, because the two differ in height and because a change occupying the third row of one may sit in the fifth row of the other. What they share is an ordered vertical structure, the setting dynamic time warping was designed for [20]. The hard minimum is not differentiable, so the soft relaxation [21] is used. Writing Delta for the pairwise cost between F and G:

$$DTW_g amma(F, G) = -gammalogsum_A exp(- < A, Delta(F, G) > /gamma) (13)$$

$$D(I, P_c^{(k)}) = DTW_g amma\left(phi(I(D)), phi(P_c^{(k)})\right) / M (14)$$

Dividing by the sentence count M turns the score into a mean per-aligned-row cost, which keeps distances comparable across length buckets. M is the document length rather than the alignment path length or the mean of the two sequence lengths because within a bucket every document shares the same M while archetypes share a fixed $M_p$, so the divisor is a batch-level constant that leaves the ranking over classes untouched and only sets the scale of the loss. The distance to a class is that to its nearest archetype, and the classifier is a softmax over negated distances with a learned scale:

$$D(I,c) = min_k D\left(I, P_c^{(k)}\right), p(y = c|I) = softmax_c\left(-D(I,c) * exp(s)\right) \quad (15)$$

The optimal alignment path is retained during analysis. It states which rows of a document correspond to which rows of its archetype, and every explanatory output below is drawn along that path.

### G. Typed Difference Maps

With the alignment fixed, the document and its nearest archetype are compared channel by channel at each row, giving one map per factor:

$$\Delta_H(r,j) = \left|\left|H\left(I(D)\right)[pi * (r), j] - H(P_c)[r,j]\right|\right| \quad (16)$$

$$\Delta_S(r,j) = \left|S\left(I(D)\right)[pi * (r), j] - S(P_c)[r,j]\right| \quad (17)$$

$$\Delta_V(r,j) = \left|V\left(I(D)\right)[pi * (r), j] - V(P_c)[r,j]\right| \quad (18)$$

Read together, the maps separate two questions a single attribution score conflates: where the disagreement sits, and which factor it belongs to. A document can match its archetype in topic throughout while diverging in sentiment across a few rows.

### H. Generative Head

The decoder g used in the Skip-Gram objective already maps pixels back toward vocabulary. A generative head psi reuses that direction and produces a distribution over tokens for any pixel vector:

$$p(w|cell) = softmax\left(psi(p)\right), psi: R^4 \to R^V \quad (19)$$

Applied to an archetype, psi answers a question no vector prototype can: what words the class representative consists of. Applied to cells where a document diverges, it shows which tokens the model expected instead.

### I. Training Objective

Seven terms make up the objective. The classification term is the negative log-likelihood of the distance-based distribution in Eq. 15, and the archetype term pulls each document toward its class representatives while a margin term pushes distinct archetypes apart:

$$L_c ls = -sum_n logp\left(y_n \middle| I(D_n)\right) \quad (20)$$

$$L_p roto = sum_n min_k D\left(I(D_n), P_{y_n}^{(k)}\right) \quad (21)$$

$$L_s ep = mean_{c!=c'} max\left(0, margin - d(P_c, P_{c'})\right) \quad (22)$$

The generative term requires archetypes to stay decodable. The auxiliary term applies channel supervision to documents and archetypes alike, so archetypes inherit the channel semantics rather than acquiring their own. The anchor term keeps archetypes near their seeds:

$$L_g en = -sum_c sum_{winclassc} logpsi(P_c)[w] \quad (23)$$

$$L_a ux = L_t opic + L_s ent, L_p aux = (L_t opic + L_s ent) overarchetypes \quad (24)$$

$$L_a nchor = sum_{c,k}\left|\left|P_c^{(k)} - P_c^{(k,init)}\right|\right|^2 \quad (25)$$

$$L = L_c ls + lambda_1 L_S G + lambda_2 L_p roto + lambda_3 L_s ep + lambda_4 L_g en + lambda_5\left(L_a ux + L_p aux\right) + lambda_6 L_a nchor \quad (26)$$

### J. Training Procedure

**1: # Stage 1 - distributional pretraining (no labels)**

2: for each epoch, for each batch of token positions:

3: compute pixels $p_{ij}$ by Eq. (2)-(4); update E, $f_{theta}$, g, psi by $L_{SG}$, Eq. (5)

**4: # Stage 2 - archetype seeding**

5: render $I(D)$ for a sample of training documents, Eq. (10)

6: for each class c: cluster its images into K groups after PCA

7: set $P_c^{(k)}$ ← row-resampled centre of cluster k

**8: # Stage 3 - joint optimization**

9: for each epoch, for each length bucket, for each batch:

10: build $I(D)$ with learned boundary rows, Eq. (8)-(10)

11: align F to each archetype by soft-DTW, Eq. (13)-(14)

12: compute L by Eq. (26); update all parameters, P and $w_B$ included

**13: # Inference**

14: $y_{ha}$ ← $argmin_c min_k D\left(I(D), P_c^{(k)}\right)$

15: return $y_{hat}$, alignment path $pi^*$, maps $\Delta_H$, $\Delta_S$, $\Delta_V$

*Algorithm 1. ProtoSemImage training and inference*

***Require**: corpus D, labels y, archetypes per class K, weights $lambda_1$*

***Ensure**: archetype bank P, prediction $y_{hat}$, alignment path, difference maps*

## IV. EXPERIMENTAL SETUP AND RESULTS

### A. Datasets

Three benchmarks cover the factors the model claims to separate. Our Category Reviews benchmark (CatRev) is the primary testbed, because its documents carry a topic and a sentiment label at the same time, and a model that entangles the two pays for it here. Five product categories supply the topic label and the accompanying star rating supplies the polarity, which reproduces the topic and sentiment structure the benchmark requires while resting on labels that were collected independently of each other. 20 Newsgroups provides topic labels only [24], which isolates the hue channel by setting the sentiment weight to zero. IMDB supplies binary sentiment [25], isolating saturation in the same way. A fourth benchmark is constructed rather than collected, because CatRev cannot test the claim that matters most to us: its labels do not depend on arrangement, so permuting sentences changes nothing a model needs to notice. CatRev-Order supplies the missing test. A document from one topic is placed before a document from another and the label is the topic of the first segment; its mirror holds the same sentences in the opposite order under the other label, so a pair shares a bag of words completely and differs only in arrangement. Table 1 lists the four benchmarks and their statistics. A fourth diagnostic is derived rather than collected: permuting sentence order inside each CatRev document while leaving the bag of words untouched. Accuracy on that permuted split measures how much of a decision depends on arrangement rather than vocabulary, and it is the test that separates an image prototype from a vector one by construction. Table 1 lists the dataset statistics.

*Table 1. Dataset Statistics*

| Dataset | Documents | Classes | Labels | Role |
|---|---|---|---|---|
| **CatRev (Amazon, five categories)** | 47,356 | 5 x 2 | 2 | Joint topic and sentiment |
| **20 Newsgroups** | 16,719 | 20 | 1 | Topic only |
| **IMDB** | 49,519 | 2 | 1 | Sentiment only |
| **CatRev-Order** | 24,000 | 5 | 1 | Order-sensitive diagnostic |

The review corpus is built to the same two-label structure as the MLR benchmark of [9], five topics crossed with two polarities, but it is not the same collection: the documents come from five Amazon product categories with the star rating supplying polarity. Accuracy figures therefore cannot be read across between the two. Two properties of these corpora constrain the design. Reviews are short, so the median CatRev document holds five sentences and the image is correspondingly shallow. Sentence counts also vary widely, from two to sixteen, which is why the encoder emits one feature per sentence and the alignment step absorbs the remaining length difference instead of padding it away.

## B. Baselines and Protocol

Every model in the main comparison is trained here, on identical splits, under an identical budget, so the numbers are directly comparable to each other. TextCNN [13] applies one-dimensional convolutions with three filter widths. A hierarchical attention network [22] runs a bidirectional GRU over tokens and then over sentences with attention at both levels. The critical baseline is a vector-prototype model that shares the embedding table, the ColorMapper, the grouped encoder, the boundary construction and the three-stage schedule with ProtoSemImage, and differs only in representing a class by a point rather than an image. Any gap between those two rows is attributable to the prototype representation, which is the claim under test. Figures published for other systems on similar benchmarks come from larger corpora and longer schedules, so they are context rather than a baseline and are not tabulated here.

## C. Implementation

The encoder is a four-block convolutional stack, the first block grouped with four groups, taking the four-channel image as input. Sentence width is 32 tokens. Archetype height is 15 rows, corresponding to eight canonical sentences. The ColorMapper is a two-layer perceptron with hidden width 256, and the generative head is a linear projection to the vocabulary. Optimization uses Adam with batch size 32, at a learning rate of 2e-4 for the Skip-Gram stage and 1e-3 for the joint stage; the second needs the larger step because at the pretraining rate the distance-based classifier is still near chance when the schedule ends. The objective weights of Eq. 26 are $\lambda_1 = 0.05$ for the Skip-Gram term, $\lambda_2 = 0.1$ for the archetype pull, $\lambda_3 = 0.05$ for the separation margin, $\lambda_4 = 0.2$ for the generative term, $\lambda_5 = 0.3$ for channel supervision on documents and archetypes alike, and $\lambda_6 = 0.01$ for the anchor penalty; the separation margin itself is 0.5. Documents are bucketed by sentence count so that no batch mixes lengths. Runs draw a 10,000-document training sample per dataset over six joint epochs, preceded by one Skip-Gram pretraining epoch. Experiments run on a single NVIDIA RTX 2080 Super. The training sample is capped well below each corpus. The cap was set by the cost of the soft-DTW recursion, which dominates runtime, and it applies equally to every row of Table 2, so the comparison remains internally valid even though absolute accuracy sits below what a longer schedule would reach.

*Table 2. Model Comparison*

| Model | CatRev (10 joint classes) | 20 Newsgroups | IMDB |
|---|---|---|---|
| **TextCNN [13]** | 61.8% | 54.8% | 84.0% |
| **HAN [22]** | 64.7% | 57.2% | 84.9% |
| **Vector prototypes (ours)** | 39.1% | 17.0% | 68.9% |
| **ProtoSemImage (ours)** | 41.5% | 17.8% | 69.8% |

## D. Results

The central comparison repeats over 3 seeds. Image prototypes average 0.3538 with a standard deviation of 0.0587, vector prototypes 0.2778 with 0.0460, and the image model is ahead in every seed, by between 4.3 and 11.8 points. Single runs move by several points, so the paired comparison rather than any one number carries the claim. Because the two models share every component except the shape of the prototype, that gap is what the image representation contributes on a task where two factors must be held apart at once. Against the sequence baselines the model trails by 23.2 points, which needs stating plainly rather than worked around.

On 20 Newsgroups the picture is worse for both. The image prototype reaches 17.8% and the vector prototype 17.0%, so the image still leads, by 0.8 points, but the pair sits far below 57.2% for the better sequence baseline. Twenty-way topic classification is where the four-channel bottleneck does the most damage, which is consistent with the diagnosis above: the more classes there are to tell apart, the more lexical detail the bottleneck has to throw away.

IMDB repeats the ordering with a smaller gap, 69.8% against 68.9%, while the sequence baselines reach 84.9%. Binary polarity is the easiest of the three tasks for the bottleneck, since one bit of supervision per sentence survives the compression better than twenty topics do. Table 3 separates two explanations for the accuracy gap. Every model in that table pushes tokens through the same four-channel bottleneck and differs only in what sits on top of it, so the softmax row measures what the bottleneck costs on its own. That diagnostic reads the encoder output by global average pooling over the sentence axis and maps the pooled vector to classes with a single linear layer, so its readout differs from the prototype models only in being pooled rather than aligned. On CatRev the softmax variant reaches 62.1% against 61.8% for TextCNN, a difference small enough to call the bottleneck exonerated on that benchmark. The image prototypes reach 41.5% from the same representation, so the shortfall sits in the distance-based matching rather than in the color compression. Twenty-way topic classification is the harder case for the compression, 44.5% against 54.8%, so there the bottleneck does cost real accuracy before matching is even considered. IMDB points the same way as CatRev, with 85.1% for the softmax variant and 84.0% for TextCNN.

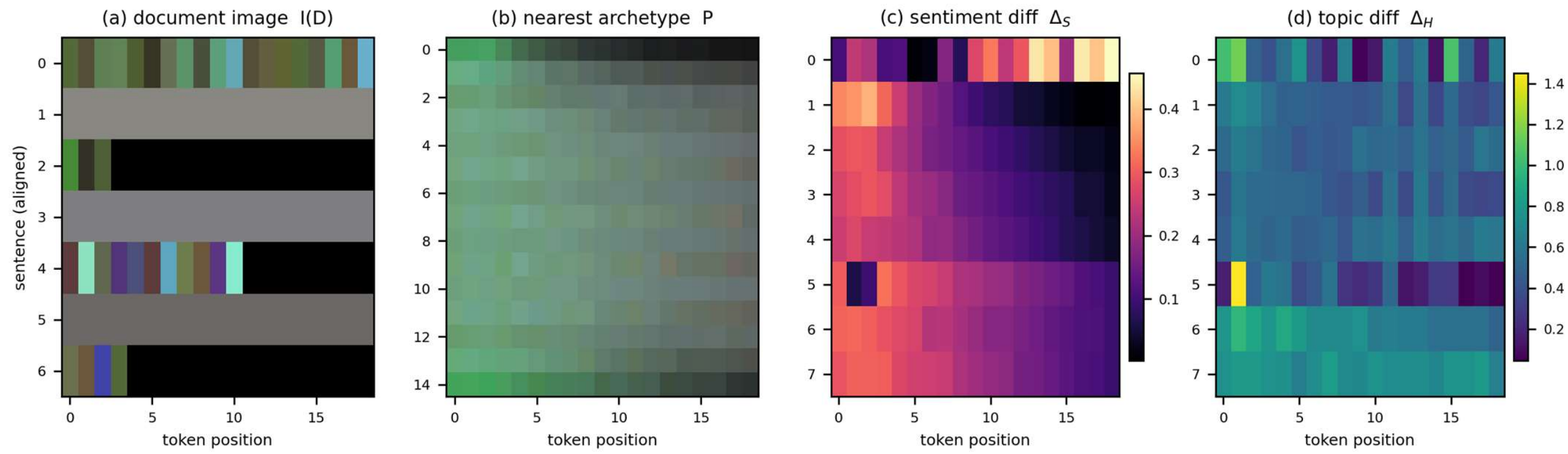

*Figure 2. Alignment and typed explanation for a held-out document. (a) the document image, (b) its nearest, (c) sentiment difference along the alignment path, (d) topic difference over the same path.*

*Table 3. Diagnostic: Same Color Bottleneck with a Direct Softmax*

| Model | CatRev | 20 Newsgroups | IMDB |
|---|---|---|---|
| Bottleneck + softmax | 62.1% | 44.5% | 85.1% |
| Bottleneck + vector prototypes | 39.1% | 17.0% | 68.9% |
| Bottleneck + image prototypes | 41.5% | 17.8% | 69.8% |
| TextCNN (full tokens) | 61.8% | 54.8% | 84.0% |

### *E. Does the Model Read Arrangement?*

Table 4 reports the order-sensitive benchmark. Every model clears chance by a wide margin, so the architecture does read arrangement: HAN reaches 91.6%, ProtoSemImage 83.7%, the vector-prototype model 74.4%, and TextCNN 42.9%, against a chance level of 20% that a bag-of-words model cannot exceed by construction.

*Table 4. Order-sensitive benchmark*

| Model | Accuracy |
|---|---|
| **HAN [22]** | 91.6% |
| **ProtoSemImage (ours)** | 83.7% |
| **Vector prototypes (ours)** | 74.4% |
| **TextCNN [13]** | 42.9% |
| **Chance (bag of words)** | 20.0% |

Figure 3 shows both results. The comparison the paper rests on sits in the middle of that list: the image prototype beats the vector prototype by 9.3 points on a task whose only signal is order. Those two models share the embedding table, the color space, the encoder, the loss weights and the schedule, and differ only in whether a class is represented by an image matched through row alignment or by a point matched by distance. The advantage that the feature-space comparison could only suggest is confirmed where it should be largest.

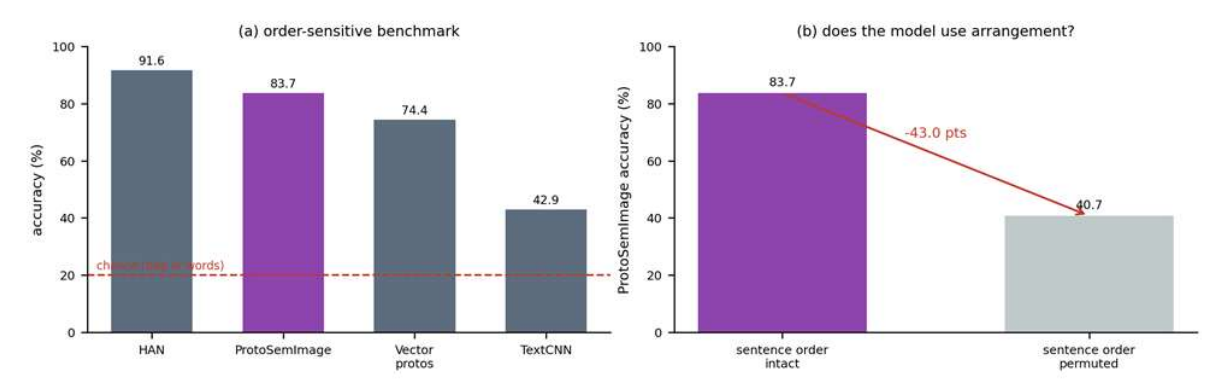

*Figure 3. (a) Accuracy on the order-sensitive benchmark, where a pair of documents shares a bag of words and differs only in arrangement. (b) Permuting sentence order inside the test documents costs ProtoSemImage 43.0 points.*

Permuting sentence order inside each test document costs ProtoSemImage 43.0 points, from 83.7% to 40.7%. On CatRev the same operation moved accuracy by 0.3 points, which we reported as inconclusive. The contrast between those numbers is the finding: the earlier null result was a property of a benchmark whose labels do not depend on order, not evidence that the representation ignores it.

## V. ABLATION STUDY

Table 5 reports the ablation on CatRev. Removing the grouped first convolution costs the most, at 13.8 points, which identifies it as the component doing the real work in this configuration. The remaining deltas are smaller, and the pattern across the row tells us which parts of the design are load-bearing on this benchmark rather than merely present.

*Table 5. Ablation Results on CatRev5. Deltas Are Relative to The Full Model. Single Runs;*

| Configuration | CatRev accuracy | Delta |
|---|---|---|
| **Full ProtoSemImage (K = 2)** | 41.5% | - |
| **K = 1 (single archetype)** | 34.6% | -6.9 pp |
| **K = 4 (more archetypes)** | 22.0% | -19.6 pp |
| **K = 4 + diversity** | 34.5% | -7.0 pp |
| **K = 4 + diversity (strong)** | 37.9% | -3.7 pp |
| **K = 8 (more archetypes)** | 15.2% | -26.4 pp |
| **K = 8 + diversity** | 23.6% | -17.9 pp |
| **Without generative head** | 33.5% | -8.0 pp |
| **Dense first convolution** | 27.8% | -13.8 pp |
| **Without Skip-Gram pretraining** | 37.6% | -3.9 pp |
| **Without auxiliary channel losses** | 28.2% | -13.3 pp |
| **Without anchor penalty** | 41.2% | -0.4 pp |

Table 5 reports the ablation. Removing the grouped first convolution costs the most, at 13.8 points, which identifies it as the component doing the real work in this configuration; the remaining deltas are smaller, and the pattern tells us which parts of the design are load-bearing here rather than merely present. Raising the number of archetypes past two dissolves the class representation rather than enriching it, and it now has a partial remedy. With more representatives available every document finds one close enough that the alignment term is satisfied trivially, and accuracy falls toward chance as K grows. The cause is that nothing keeps the archetypes of one class apart. Adding an intra-class diversity term, which penalizes close pairs within a class rather than only between classes, recovers most of the loss: K = 4 rises from 22.0% to 37.9%, and K = 8 from 15.2% to 23.6%. Two archetypes per class remains the best setting, but the collapse is now a diagnosed and partly treatable effect rather than an unexplained one.

## VI. ANALYSIS AND DISCUSSION

### A. *Reading an Archetype*

The interpretability claim rests on archetypes being displayable, so it is worth being precise about what a reader sees. An archetype image has the same structure as a document image: one row per canonical sentence, boundary rows between them, hue carrying topic, saturation carrying sentiment, value carrying emphasis. A reader can observe that a class archetype holds a steady hue across its upper rows and shifts partway down, which is a statement about typical discourse progression in that class. No vector prototype supports that reading, because a point has no vertical axis.

### B. *Locating Disagreement*

Figure 2 shows the second output. A document is aligned to its nearest archetype, and the difference maps are drawn along that path. Where hue agreement is close to uniform while saturation diverges over a contiguous band of rows, the reading is that the document belongs to the archetype's topic but departs from its emotional profile in a specific region, which is more actionable than a single attribution weight.

### C. *What the Evidence Supports, and What It Does Not*

Table 6 sets the four claims that motivated the design against what the experiments showed, and they did not fare equally: 1 of the four survives contact with the experiments and 3 do not.

*Table 6. Hypotheses Against Measured Outcomes on CatRev.*

| Hypothesis | Verdict | Evidence |
|---|---|---|
| **Image prototypes beat vector prototypes** | supported | 0.3538 vs 0.2778 averaged over 3 seeds; ahead in every seed |
| **Image prototypes exploit discourse order** | supported | permuting order costs 43.0 points on CatRev-Order, and the image prototype leads the vector one by 9.3 points there |
| **The four-channel bottleneck causes the accuracy gap** | not supported | a softmax head on the same representation reaches 62.1% against 61.8% for TextCNN |
| **More archetypes per class help** | not supported | K = 1 gives 34.6%, K = 2 gives 41.5%, K = 4 gives 22.0%, K = 8 gives 15.2% |

The first row is the claim the paper can defend, and it is the reason the approach is worth pursuing: the same pipeline, the same color space and the same schedule produce a better classifier when the prototype is an image. The second row is the claim CatRev could not test, because its labels do not depend on arrangement, and which CatRev-Order settles. The model does read layout: destroying order costs it 43 points, and on that benchmark the image prototype leads the vector one by more than nine. The earlier null result was a property of the benchmark, which is the whole reason a second benchmark exists.

The third-row matters because it removes the most convenient excuse. It would have been tidy to blame the color compression for the accuracy gap and leave the matching untouched. The diagnostic says otherwise: a direct softmax head on the identical representation reaches the sequence baselines, so the compression is largely exonerated and the distance-based matching is where the loss sits. That is a harder finding to write around, and it points at the matching as the component that needs work.

The fourth row is the clearest failure. Adding archetypes beyond two does not enrich the class representation, it dissolves it. With enough representatives every document finds one close enough that the alignment objective is satisfied without the model learning to discriminate, and accuracy falls toward chance as K grows. Two archetypes per class, held in place by the anchor penalty, is the configuration that avoids the collapse.

## VII. LIMITATIONS AND OPEN QUESTIONS

Four constraints bound everything reported above, and stating them together is more useful than distributing them through the results. The training budget is the first. Every controlled run draws a 10,000-document sample and trains for six joint epochs, a cap set by the cost of the soft-DTW recursion rather than by what the models need. The cap applies equally to every row of the comparison, so the ordering between models should survive, but the absolute numbers are floors rather than ceilings, and the gap to the sequence baselines may narrow with a longer schedule. A reader should treat the ranking as the finding and the magnitudes as provisional.

Layout is now tested rather than assumed, but on a benchmark we built. CatRev-Order isolates arrangement by holding the bag of words fixed, which makes it decisive as a diagnostic and artificial as a task: real documents rarely differ only in the order of their halves. Whether the same advantage appears on naturally order-dependent problems, such as argument structure or narrative flow, is open, and it is the next experiment we would run. A second caveat is that HAN still leads on CatRev-Order, by 7.9 points, so the image prototype reads arrangement better than its vector counterpart without reading it better than a dedicated sequential model. The distance-based matching is undertrained rather than fundamentally inadequate, as far as we can tell. It has one hyperparameter that matters, the number of archetypes per class, and the response to it is not smooth: one archetype is too few, two works, and four or more collapses toward chance because documents find a representative close enough that the alignment objective is satisfied without the model learning to discriminate. A mechanism that fragile is a mechanism we do not yet understand well, and the collapse is worth studying on its own terms rather than only as an ablation row.

Finally, the review corpus is not the corpus used by the earlier work we compare against. It shares the two-label structure, five topics crossed with two polarities, but the documents come from a different source, so accuracy figures published for other systems cannot be read as a baseline for these runs.

## VIII. CONCLUSION AND FUTURE WORK

ProtoSemImage asked whether a class prototype can be an image, and the answer is yes with a caveat that matters. Archetypes live in the same four-channel HSV space as the documents they represent, which lets a decision be made by deformable row alignment and explained by channel-wise difference maps along the resulting path, and a generative head renders them as readable token distributions. On the benchmark where two factors have to be held apart at once, image prototypes beat vector prototypes in every seed we ran. The distance-based matching that makes the representation interpretable is also what keeps the model behind the sequence baselines, and the diagnostic that separates those two effects is, we think, the most useful single result here: it rules out the explanation that would have been most convenient.

The next step is not a new architecture but a better classifier on top of the existing one. Cross-attention between document rows and archetype rows would replace the dynamic programming recursion with a learned alignment, keeping the spatial and channel-wise comparison that makes the explanation readable while giving the matching the capacity it currently lacks. Attention has already been integrated into image-to-image models to good effect [27], which makes the same substitution worth trying here. An order-sensitive benchmark, built so that the same bag of sentences carries different labels depending on arrangement, would give the layout question the test it deserves. Beyond that, archetypes could be communicated between parties instead of model parameters, since they are small and carry no source text, which raises the question of what they reveal and what they conceal. The bank could grow as new classes appear, turning the distance threshold into a mechanism for admitting classes rather than only for rejecting documents. Replacing the convolutional encoder with a vision transformer would test whether row alignment still supplies most of the structure once global attention is available. And the generative head invites a harder question than reconstruction: whether a document synthesized from an archetype can serve as a training example in its own right.